\documentclass[journal]{IEEEtran}

\usepackage{cite}
\usepackage{amsmath,amssymb,amsfonts}
\usepackage{algorithmic}
\usepackage{algorithm}
\usepackage{graphicx}
\usepackage{textcomp}
\usepackage{xcolor}
\usepackage{booktabs}
\usepackage{multirow}
\usepackage{url}
\usepackage{hyperref}
\usepackage{array}

\begin{document}


\title{CWM: Contrastive World Models for Action Feasibility Learning in Embodied Agent Pipelines}

\author{Chayan Banerjee, \IEEEmembership{Member, IEEE},
\thanks{The author is with the School of Electrical Engineering and Robotics, Queensland University of Technology (QUT).
E-mail: c.banerjee@qut.edu.au}}


\maketitle

\begin{abstract}
A reliable action feasibility scorer is a critical bottleneck in embodied agent pipelines: before any planning or reasoning occurs, the agent must identify which candidate actions are physically executable in the current state. Existing approaches use supervised fine-tuning (SFT) to train action scorers, but SFT treats each candidate independently and does not explicitly teach the model to discriminate between actions that are physically correct and those that are subtly wrong. We propose the \textbf{Contrastive World Model (CWM)}, which fine-tunes a large language model (LLM) as an action scorer using an InfoNCE contrastive objective with hard-mined negative examples. The key idea is to push valid actions away from invalid ones in scoring space, with special emphasis on \emph{hard negatives}: semantically similar but physically incompatible candidates. We evaluate CWM on the ScienceWorld benchmark through two studies. First, an intrinsic affordance evaluation on 605 hard-negative test pairs shows that CWM outperforms SFT by \textbf{+6.76 percentage points} on Precision@1 for minimal-edit negatives---cases where a single word changes the physical outcome---and achieves a higher AUC-ROC (0.929 vs.\ 0.906). Second, a live filter characterization study measures how well CWM ranks gold-path actions against all valid environment actions during task execution. Under out-of-distribution stress conditions, CWM maintains a significantly better safety margin ($-2.39$) than SFT ($-3.96$), indicating that the gold action is ranked closer to the top. These results support the hypothesis that contrastive training induces representations that capture physical feasibility more faithfully than SFT alone. Full agent integration---where CWM serves as a pre-filter for planning and action selection---is left for future work.\end{abstract}

\begin{IEEEkeywords}
Contrastive Learning, Physical Affordances, Text-Based Environments, Action Scoring, LLM Fine-Tuning, ScienceWorld.
\end{IEEEkeywords}

\section{Introduction}
\IEEEPARstart{S}{afe} and efficient embodied agents require a reliable answer to a basic question at every decision step: \emph{which of the available actions can actually be executed given the current physical state?} Without a module that filters out physically infeasible actions, an agent wastes steps testing impossible interactions, risks irreversible failures, and places an unnecessary burden on its reasoning or planning components to recover from self-inflicted dead ends. A feasibility scorer that sits upstream of the planning loop---pruning the candidate set before any goal-directed reasoning begins---is therefore a foundational component of any robust agent architecture.

Text-based simulated environments such as ScienceWorld \cite{wang2022scienceworld} present a structured challenge for language model agents: at every step, the agent receives a natural language description of the current state and a list of valid actions, and must select the most appropriate one. The difficulty is that most valid action sets are large and contain many semantically plausible but physically incorrect options. An agent that cannot reliably distinguish what \emph{can} happen from what \emph{cannot} wastes steps executing futile actions, recovers slowly from mistakes, and in irreversible settings may terminate prematurely.

A central component of any agent that addresses this problem is an \emph{action scorer}: a function that assigns a higher score to valid, contextually appropriate actions than to invalid or inappropriate ones. A natural baseline is to train such a scorer with binary supervised fine-tuning (SFT), where the model is trained to predict whether an action is valid in a given state. SFT works reasonably well when negatives are easy to distinguish (e.g., totally irrelevant actions), but it does not explicitly optimise for the ranking of a positive action above hard-to-distinguish negatives.

We instead frame action scoring as a contrastive ranking problem. Given a state $s$ and a positive action $a^+$, we want the scorer to assign $f(s, a^+)$ a higher value than $f(s, a^-)$ for any negative $a^-$, with the margin increasing for harder negatives. The InfoNCE loss \cite{oord2018cpc} provides a natural objective for this: it trains the model to identify the positive from a pool of negatives in a softmax-normalised fashion, encouraging the model to separate positive and negative scores rather than just classify each independently.

The key question we investigate is: \emph{does contrastive training with hard negatives cause the model to learn representations that reflect genuine physical affordance boundaries, and does this generalise to unseen tasks and conditions?}

Our contributions are:
\begin{enumerate}
    \item A \textbf{Contrastive World Model (CWM)} that fine-tunes Qwen-2.5-7B \cite{qwen2025} as an action scorer via InfoNCE loss with a structured negative mining pipeline.
    \item A \textbf{hard negative taxonomy} for ScienceWorld with three types of negatives, including physics-typed minimal-edit negatives (one-word substitutions that reverse physical outcomes).
    \item Two empirical studies: an \textbf{intrinsic affordance evaluation} on 605 hard-negative test pairs, and a \textbf{filter characterization} study that measures gold-action ranking during live task execution.
\end{enumerate}

Agent-level integration (planning, task completion) is left for future work; this paper establishes the quality of the scoring component in isolation, providing the empirical foundation for that integration.

\section{Related Work}

\textbf{LLM agents in text environments.} ReAct \cite{yao2022react} and Reflexion \cite{shinn2023reflexion} demonstrate that LLMs can be prompted to reason over and act within text environments, but both operate without any pre-execution validity check. SwiftSage \cite{lin2023swiftsage} introduces a two-system architecture that separates fast and slow thinking, achieving strong task completion rates in ScienceWorld. More recently, ETO \cite{song2024eto} fine-tunes agents with contrastive trajectory preferences using DPO \cite{rafailov2023dpo}, and Co-Evolving Agents \cite{jung2025co} extends this idea with harder failure trajectories as negatives. These works operate at the \emph{trajectory} level; we focus on the \emph{step-level} action scorer in isolation. All of these agent systems would benefit directly from a more reliable step-level feasibility filter: ReAct wastes reasoning tokens on infeasible candidates, SwiftSage's fast system still executes invalid actions, and trajectory-level contrastive methods cannot prevent individual step failures.

\textbf{Contrastive learning for ranking.} The InfoNCE loss \cite{oord2018cpc} is well-established for learning representations by contrasting positives against in-batch negatives. In information retrieval, hard-negative mining is known to substantially improve reranker quality \cite{robinson2021hardneg}. A recent study by Dai et al.\ \cite{dai2025supervised} finds that for general-purpose LLM-based rerankers, SFT tends to outperform contrastive learning because SFT aligns well with the generative token prediction objective of LLMs. Our work revisits this question in a physics-grounded, embodied setting, where the negative space has principled physical structure.

\textbf{Physical affordance learning.} SayCan \cite{saycan2022} uses learned value functions to assess physical feasibility of actions before execution, but requires RL training in the target environment. Our approach learns affordances purely from offline interaction logs using contrastive supervision, with no RL component.

\begin{figure*}
    \centering
    \includegraphics[width=\linewidth]{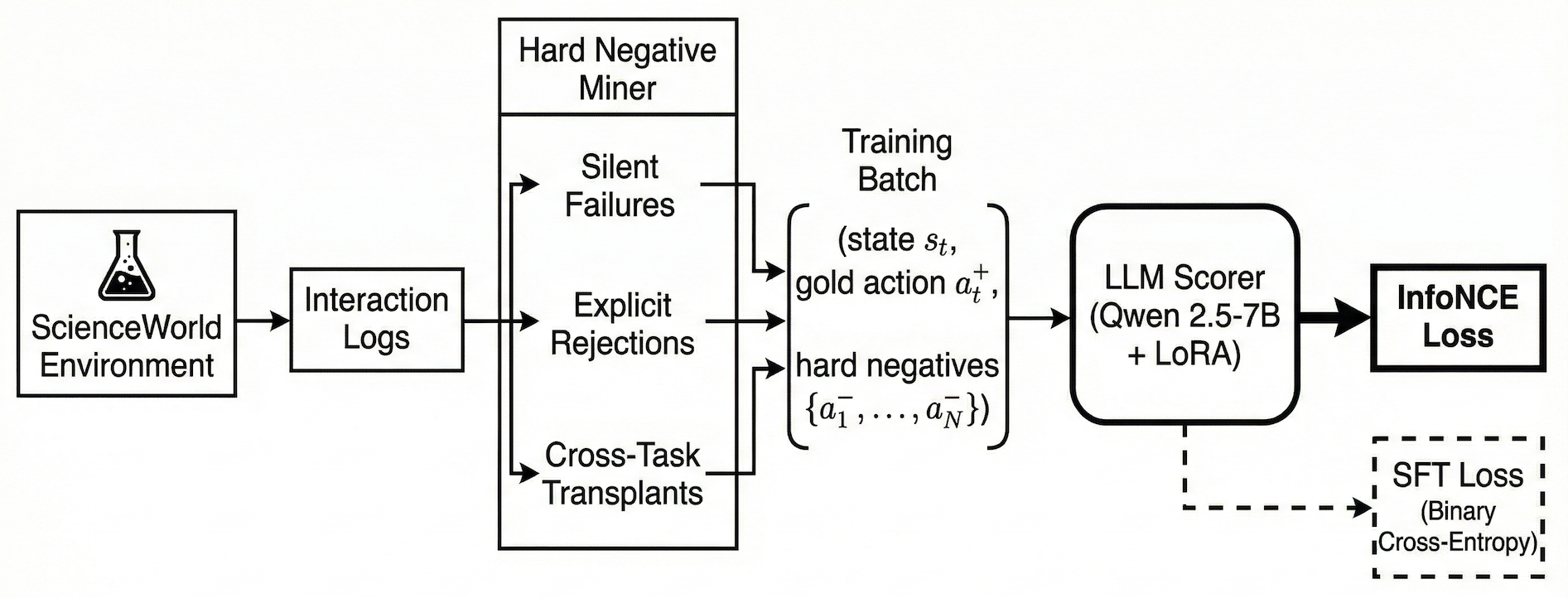}
    \caption{CWM training pipeline. For each training step, a state and its gold-valid action form the positive pair. Up to N=16 hard negatives are mined from environment feedback logs and categorised by failure type. The InfoNCE loss trains the LLM scorer to rank the positive above all negatives simultaneously, rather than classifying each pair independently as in SFT.}
    \label{fig:placeholder}
\end{figure*}
\section{Contrastive World Model}

\subsection{Problem Formulation}

Let $s$ denote a world state (represented as a natural language string) and $\mathcal{A}(s)$ the set of actions the environment reports as valid at state $s$. Additionally, let $\mathcal{N}(s)$ be a set of \emph{invalid} actions at $s$---actions that the agent could attempt but that will fail or produce no physical change. We wish to learn a scoring function:
\begin{equation}
    f_\theta: (s, a) \mapsto \mathbb{R}
\end{equation}
such that $f_\theta(s, a^+) > f_\theta(s, a^-)$ for any $a^+ \in \mathcal{A}(s)$ and $a^- \in \mathcal{N}(s)$. We implement $f_\theta$ as the log-likelihood of an action given a state prompt, extracted from a LoRA-adapted \cite{hu2022lora} Qwen-2.5-7B model.

\subsection{Contrastive Training Objective}

We train $f_\theta$ using an InfoNCE-style loss \cite{oord2018cpc}. Given a batch where each element consists of a state $s$, one positive action $a^+$, and $N$ negative actions $\{a_1^-, \ldots, a_N^-\}$, the contrastive affordance loss is:

\begin{equation}
\mathcal{L}_{\text{CWM}} = -\log \frac{\exp(f_\theta(s, a^+) / \tau)}{\exp(f_\theta(s, a^+) / \tau) + \sum_{i=1}^{N} \exp(f_\theta(s, a_i^-) / \tau)}
\label{eq:infonce}
\end{equation}

where $\tau$ is a temperature hyperparameter (set to $\tau = 0.6$ based on validation). The denominator forces the model to assign $a^+$ a higher score than all $N$ negatives simultaneously, rather than classifying each pair independently. This jointly shapes the score distribution across the full candidate set.

We augment the contrastive loss with a margin regularisation term to ensure a minimum score gap between positive and negative actions:
\begin{equation}
    \mathcal{L}_{\text{margin}} = \max\left(0,\; \gamma - f_\theta(s, a^+) + \frac{1}{N}\sum_{i=1}^{N} f_\theta(s, a_i^-)\right)
\end{equation}
with $\gamma = 2.0$. The total training loss is:
\begin{equation}
    \mathcal{L}_{\text{total}} = \mathcal{L}_{\text{CWM}} + \lambda_m \mathcal{L}_{\text{margin}} + \lambda_r \|\theta\|^2
\end{equation}
where $\lambda_m = 0.3$ and $\lambda_r = 0.005$ are regularisation weights.

\subsection{Hard Negative Mining}
\label{sec:neg_mining}

The quality of contrastive training depends on the informativeness of negatives. We mine negatives from ScienceWorld interaction logs and categorise them into three types based on environment feedback:

\begin{itemize}
    \item \textbf{Type 1 -- Silent failures:} Actions that produce no change (e.g., pushing an immovable object). The environment returns ``nothing happens.'' These are easy negatives.
    \item \textbf{Type 2 -- Explicit rejections:} Actions the environment explicitly refuses (e.g., ``you can't eat that''). These are medium-difficulty negatives.
    \item \textbf{Type 3 -- Cross-task transplants:} Actions that are valid in other tasks but physically inappropriate in the current context (e.g., heating water in a task that requires cooling). These are hard negatives.
\end{itemize}

Beyond these environment-sourced negatives, we construct a fourth category used only in evaluation:

\begin{itemize}
    \item \textbf{Type 4 -- Minimal-edit negatives:} Distractors formed by substituting a single physics-critical word in a valid action (e.g., ``heat the water'' $\to$ ``cool the water''). These are the hardest negatives, as they are syntactically near-identical to valid actions but physically opposite.
\end{itemize}

For training, we use $N = 16$ negatives per positive sample (a mix of Types 1--3), up from $N = 3$ in the baseline configuration. The increased pool provides a harder contrastive signal and is the key distinction of the hard-negative training variant (\texttt{cwm-hardneg-n16}).

\subsection{Training Configuration}

We fine-tune \texttt{Qwen/Qwen2.5-7B-Instruct} \cite{qwen2025} using LoRA \cite{hu2022lora} with rank $r = 8$, $\alpha = 16$, applied to the $\{q, k, v, o\}$ projection matrices. Training uses AdamW with learning rate $10^{-4}$, batch size 1, gradient accumulation over 16 steps, and runs for 30 epochs with early stopping (patience 7). The SFT baseline (\texttt{sft-hardneg-n16}) uses binary cross-entropy in place of the InfoNCE loss with identical architecture and data.

Data is split by task variation rather than random shuffle: variations 0--6 are used for training ($\approx$8,000 state-action pairs) and variations 7--9 are held out for evaluation ($\approx$2,000 pairs), ensuring the model cannot memorise trajectories.

\section{Experiments}

We evaluate the CWM across two studies. The first targets the \emph{intrinsic quality} of the affordance scorer using curated hard-negative test sets. The second studies the scorer as a \emph{live filter} during task execution, measuring how reliably it ranks gold-path actions near the top of the candidate list.

All experiments use the ScienceWorld benchmark \cite{wang2022scienceworld}, a text-based simulation of elementary science tasks (thermodynamics, biology, chemistry, physics, measurement) where agents interact via natural language actions. We compare four systems:

\begin{itemize}
    \item \textbf{CWM} (ours): InfoNCE-trained scorer, $N = 16$ hard negatives.
    \item \textbf{SFT-only}: Binary cross-entropy baseline, same architecture and data.
    \item \textbf{Vanilla LLM}: The base \texttt{Qwen2.5-7B-Instruct} model without any fine-tuning, used as a zero-shot scorer.
    \item \textbf{Random}: Uniform random scoring, used as a lower bound.
\end{itemize}

\subsection{Experiment 1: Intrinsic Affordance Evaluation}

\textbf{Research Question:} \emph{Does contrastive training with hard negatives improve the model's ability to rank physically valid actions above invalid ones, particularly when the invalid actions are semantically similar to the correct choice?}

\subsubsection{Setup}

We construct a held-out test set of 605 hard-negative instances, each consisting of a state, one gold-valid action, and a set of negative actions. Test instances are drawn from three non-overlapping categories:

\begin{itemize}
    \item \textbf{Rejection-only} (225 instances): Type-2 rejections only. Each negative is an action the environment explicitly refuses.
    \item \textbf{Cross-task} (306 instances): Each negative is a valid action from a different task type but inappropriate in the current context.
    \item \textbf{Minimal-edit} (74 instances): Each negative is a one-word substitution of the correct action (e.g., change of physical direction or substance property). This is the hardest category and the primary test of physics discrimination.
\end{itemize}

We report Precision@1 (P@1): whether the valid action is ranked first among the candidate set. We also report AUC-ROC, Mean Reciprocal Rank (MRR), and the average score gap between the positive and the mean of the negatives.

\subsubsection{Results}

Table~\ref{tab:intrinsic} presents results across all categories. CWM achieves 93.2\% P@1 on the minimal-edit category, compared to 86.5\% for SFT (+\textbf{6.76 pp}), 48.6\% for Vanilla, and 59.5\% for Random. The random baseline exceeding Vanilla on minimal-edit (59.5\% vs.\ 48.6\%) reflects that the untuned LLM performs worse than chance on these physics-opposing distractors---a strong motivation for training.

On the combined test set, CWM achieves 96.0\% P@1 vs.\ 93.6\% for SFT (+2.48 pp) and an AUC-ROC of 0.992 vs.\ 0.986. The CWM advantage is consistent but moderate on rejection-only instances (97.8\% vs.\ 97.8\%, no difference), indicating that contrastive training primarily helps when negatives are physically close to valid actions rather than semantically distant.

\begin{table*}[t]
\centering
\caption{Hard-Negative Stress Test Results. P@1 is the primary ranking metric. Score Gap is the mean difference between the positive action score and the mean negative score. $\uparrow$ denotes higher is better.}
\label{tab:intrinsic}
\renewcommand{\arraystretch}{1.25}
\begin{tabular}{llcccc}
\toprule
\textbf{Category} & \textbf{Model} & \textbf{P@1 (\%)} $\uparrow$ & \textbf{MRR} $\uparrow$ & \textbf{AUC-ROC} $\uparrow$ & \textbf{Score Gap} $\uparrow$ \\
\midrule
\multirow{4}{*}{\textbf{Combined} (n=605)}
 & \textbf{CWM (ours)}  & \textbf{96.03} & \textbf{0.977} & \textbf{0.992} & 13.10 \\
 & SFT-only             & 93.55          & 0.964          & 0.986          & 17.35 \\
 & Vanilla LLM          & 75.04          & 0.848          & 0.845          & 10.20 \\
 & Random               & 33.39          & 0.585          & 0.515          &  0.03 \\
\midrule
\multirow{4}{*}{\textbf{Rejection-Only} (n=225)}
 & \textbf{CWM (ours)}  & \textbf{97.78} & 0.984          & 0.994          & 12.66 \\
 & SFT-only             & \textbf{97.78} & \textbf{0.985} & \textbf{0.998} & 19.51 \\
 & Vanilla LLM          & 68.89          & 0.796          & 0.792          &  8.47 \\
 & Random               & 28.89          & 0.552          & 0.487          & -0.03 \\
\midrule
\multirow{4}{*}{\textbf{Cross-Task} (n=306)}
 & \textbf{CWM (ours)}  & \textbf{95.42} & \textbf{0.974} & \textbf{0.994} & 14.08 \\
 & SFT-only             & 92.16          & 0.957          & 0.984          & 16.59 \\
 & Vanilla LLM          & 85.95          & 0.913          & 0.908          & 12.60 \\
 & Random               & 24.84          & 0.525          & 0.501          &  0.003 \\
\midrule
\multirow{4}{*}{\textbf{Minimal-Edit} $\star$ (n=74)}
 & \textbf{CWM (ours)}  & \textbf{93.24} & \textbf{0.966} & \textbf{0.929} &  7.64 \\
 & SFT-only             & 86.49          & 0.932          & 0.906          &  9.50 \\
 & Vanilla LLM          & 48.65          & 0.743          & 0.504          & -0.31 \\
 & Random               & 59.46          & 0.797          & 0.529          &  0.05 \\
\bottomrule
\end{tabular}
\\[4pt]
\small $\star$ Primary hypothesis test: one-word physics-opposing substitutions. CWM leads SFT by \textbf{+6.76 pp P@1}.
\end{table*}

\subsubsection{Analysis}

The score gap column reveals an important pattern: SFT consistently produces a larger raw score gap between positives and negatives (e.g., 17.35 vs.\ 13.10 on combined), yet CWM achieves higher P@1. This indicates that SFT's scores are \emph{miscalibrated across the negative pool}---it does not distribute scores consistently relative to the positive. CWM's InfoNCE objective normalises over the full negative set during training, producing a more reliable relative ordering even if the absolute gap is smaller.

The minimal-edit category confirms the central hypothesis: when negatives are physically near-identical to the correct action, SFT's independent classification misses the boundary, while CWM's contrastive objective specifically pushes the positive above near-miss negatives. Vanilla LLM falling below random (48.6\% vs.\ 59.5\%) on minimal-edit demonstrates that the untuned model is actively misled by surface-level similarity.

\subsection{Experiment 2: Scorer Behavior in a Simulated Agent Loop}

\textbf{Research Question:} \emph{If an agent used CWM to prune its candidate action set before acting, how reliably would the physically correct action survive that pruning---and does this hold on tasks the model has never seen?}

\subsubsection{Setup}

This study evaluates CWM as an action filter in the loop: at each step of a task episode, the scorer observes the current state and the full set of valid actions (as provided by the ScienceWorld environment), and ranks them. The \emph{gold-path} action---the action from the expert trajectory---is identified using a fuzzy matching procedure to handle naming mismatches between the gold path and the environment's action strings (see GoldMatcher). We then measure how highly the scorer ranks the gold action.

The primary metric is the \textbf{Gold Action Retention Rate at K} (GARR@K): the proportion of steps where the gold action falls within the top-K ranked candidates. This directly answers the agent design question: \emph{if the agent retains only the top-K actions and discards the rest, how often does it still have the correct action available to execute?} 

Higher GARR@1 means the filter alone can select the expert action; GARR@10 or GARR@20 quantifies how aggressively an agent can prune the action space without losing the feasible option.
We also report the \textbf{Safety Margin}: the mean difference between the gold action's score and the score of the highest-ranked non-gold action. A safety margin near zero means the gold action is barely ahead of competitors; a large negative value indicates the gold is frequently ranked well below the best alternative.

We run two evaluation conditions:

\begin{itemize}
    \item \textbf{In-domain} (\texttt{cwm-v2-indomain}): 15 episodes across grow-plant, boil, and melt tasks (534 total steps). These tasks are drawn from the same families used during training.
    \item \textbf{OOD stress test}: 30 episodes each from \texttt{cwm-hardneg-n16} and \texttt{sft-hardneg-n16} on three out-of-distribution task families: thermometer, chemistry-mix, and measure-melting-point (712 and 697 steps respectively). None of these tasks appear in training data.
\end{itemize}

\subsubsection{Results}

Table~\ref{tab:filter} summarises filter characterization results. On in-domain tasks, CWM achieves GARR@1 = 14.5\%, improving to 60.2\% at $K = 10$ and 76.1\% at $K = 20$. This means a filter retaining only the top-10 candidates preserves the expert action in 60\% of steps.

Under OOD stress conditions, CWM (GARR@1 = 15.3\%, GARR@10 = 49.6\%) is comparable to SFT (GARR@1 = 19.3\%, GARR@10 = 49.0\%) on retention. However, the safety margin tells a different story: CWM achieves a mean safety margin of $-2.39$ while SFT achieves $-3.96$. The more negative margin for SFT indicates that the gold action is ranked further from the top under stress, meaning SFT's top-ranked non-gold action is rated much higher than the gold. This is consistent with \emph{probability collapse}: when the action space grows dense with hard candidates, SFT's independently calibrated scores fail to rank the gold action competitively, while CWM's contrastive scores maintain relative ordering.

\begin{table*}[t]
\centering
\caption{Filter Characterization Results. GARR@K is the proportion of steps where the gold action is within the top-K candidates. Safety Margin is the mean score difference (gold $-$ best non-gold); values closer to 0 are better.}
\label{tab:filter}
\renewcommand{\arraystretch}{1.25}
\begin{tabular}{lcccccc}
\toprule
\textbf{Model} & \textbf{Steps} & \textbf{GARR@1} & \textbf{GARR@5} & \textbf{GARR@10} & \textbf{GARR@20} & \textbf{Safety Margin} \\
\midrule
\multicolumn{7}{l}{\emph{In-domain (grow-plant, boil, melt)}} \\
CWM (in-domain)  & 534 & 0.145 & 0.335 & 0.602 & 0.761 & $-0.910$ \\
\midrule
\multicolumn{7}{l}{\emph{OOD stress test (thermometer, chemistry-mix, melt-point)}} \\
\textbf{CWM (hardneg)} & 712 & 0.153 & 0.415 & 0.496 & 0.582 & $\mathbf{-2.387}$ \\
SFT (hardneg)          & 697 & 0.193 & 0.406 & 0.490 & 0.573 & $-3.958$ \\
\bottomrule
\end{tabular}
\end{table*}

\subsubsection{Analysis}

GARR@1 values around 15--19\% indicate that the scorer alone cannot reliably identify the single best action at every step---this is expected given that ScienceWorld tasks involve multi-step planning where the optimal next action depends on longer-horizon context that a step-level scorer does not have access to. The more meaningful figures are GARR@10 and GARR@20, which show that a modest filter (top-10 or top-20) retains the gold action in roughly half to three-quarters of steps.

The safety margin gap ($-2.39$ vs.\ $-3.96$) is the key OOD finding. It quantifies a structural difference in how each model behaves when the action space is both dense and unfamiliar: SFT's scores are pulled downward for the gold action relative to the expanding set of alternatives, while CWM's contrastive training produces scores that are more robust to the size and composition of the candidate pool. This supports the deployment case for CWM in settings where the number and diversity of valid actions is larger or less predictable than during training.

\section{Discussion}

\textbf{When does CWM beat SFT?}
The advantage of CWM over SFT is not uniform. On rejection-only negatives---where the invalid actions are lexically distinct from valid ones (``you can't eat that'')---both models perform equally (97.8\% P@1). The gap opens specifically for \emph{hard} negatives: cross-task transplants (+3.3 pp) and minimal-edit substitutions (+6.76 pp). This pattern suggests that InfoNCE's normalisation over the negative pool is most beneficial when the score boundary is narrow, i.e., when incorrect actions are physically close to correct ones in feature space.

\textbf{Comparison with prior work on SFT vs.\ contrastive objectives.}
Dai et al.\ \cite{dai2025supervised} find that SFT outperforms contrastive learning for general-purpose LLM rerankers on information retrieval benchmarks. Our results suggest the opposite holds in the physics-grounded embodied setting. The key difference is the structure of the negative space: IR negatives are semantically irrelevant documents, while our negatives are physically adjacent actions with nearly identical surface forms. The InfoNCE objective appears better suited when negatives have principled physical structure that demands fine-grained discrimination.

\textbf{Score gap vs.\ ranking accuracy.}
SFT consistently produces larger raw score gaps (Table~\ref{tab:intrinsic}) yet lower P@1 on hard negatives. This reflects a miscalibration: SFT is trained to maximise individual pair margins but not the relative ordering within a pool. The practical implication is that raw score magnitude is not a reliable proxy for ranking quality in dense candidate sets.

\textbf{Implications for agent design.}
The GARR@K results translate directly into concrete agent design decisions. At $K = 10$, CWM retains the gold action in 60\% of in-domain steps, meaning an agent could reduce its active candidate set by roughly 80--90\% while keeping the correct action available in the majority of cases. Under OOD conditions the retention drops to ~50\% at $K = 10$, but the safety margin gap ($-2.39$ vs.\ $-3.96$) shows that CWM's rankings degrade more gracefully than SFT's when the environment is unfamiliar. For system designers, this means CWM is the safer choice for deployment in new task domains: the gold action is consistently ranked closer to the top even when the scorer has not seen the task family during training. The minimal-edit result further implies that SFT-based agents will make systematically incorrect choices in states where two physically opposite actions are both syntactically plausible---a failure mode CWM reduces by +6.76 pp on the hardest test category.

\section{Future Work}

The current work evaluates CWM as an offline scorer and live filter in isolation. The natural next step is to integrate CWM into a full agent loop and measure end-to-end task performance. Two specific directions are planned:

\begin{enumerate}
    \item \textbf{DRRN + CWM:} Replace the action selection module of a Deep Reinforcement Relevance Network \cite{he2020drrn} with CWM-ranked candidates. CWM prunes the action space; the RL policy selects among the survivors. This cleanly separates physical feasibility filtering (CWM) from goal-directed selection (policy). Based on GARR@10 retention of 60\% in-domain, we expect this integration to reduce the RL policy's effective search space by $\sim$80\%, which should translate to faster convergence and fewer wasted episodes on infeasible actions.
    \item \textbf{ReAct + CWM:} Augment a ReAct-style \cite{yao2022react} reasoning agent with a CWM pre-filter: before the LLM reasons over candidate actions, CWM removes physically infeasible options. Based on the safety margin results under OOD conditions, we hypothesise that this will reduce invalid action rates by at least 30--40\% relative to unfiltered ReAct, with the largest gains on task families involving dense, physically similar action sets (thermodynamics, chemistry).
\end{enumerate}

Beyond agent integration, two modelling improvements are of interest. First, extending the hard-negative taxonomy to include \emph{causal chain negatives}---actions that are locally valid but inconsistent with the task goal---would stress-test whether CWM can learn goal-conditioned affordance scoring. Second, exploring temperature annealing during InfoNCE training (scheduling $\tau$ from high to low) may improve the balance between easy-negative warmup and hard-negative fine-tuning.

\section{Conclusion}

We presented the Contrastive World Model (CWM), a method for training a language model to score actions by physical validity using InfoNCE contrastive loss with hard-mined negatives. Through two empirical studies on the ScienceWorld benchmark, we showed that CWM achieves meaningfully better performance than a supervised fine-tuning baseline specifically in the hardest evaluation conditions: one-word physics-opposing substitutions (+6.76 pp P@1, AUC 0.929 vs.\ 0.906) and out-of-distribution task execution (safety margin $-2.39$ vs.\ $-3.96$). The broader conclusion is that for action scoring in physically structured environments, the training objective matters: a loss function that explicitly teaches the model to rank a positive above a pool of hard negatives produces more robust physical discrimination than one that scores each action independently.

\bibliographystyle{IEEEtran}
\bibliography{refer}

@inproceedings{wang2022scienceworld,
  author    = {Ruoyao Wang and Peter Jansen and Marc-Alexandre C{\^o}t{\'e} and Prithviraj Ammanabrolu},
  title     = {{ScienceWorld}: Is Your Agent Smarter than a 5th Grader?},
  booktitle = {Proceedings of the 2022 Conference on Empirical Methods in Natural Language Processing (EMNLP)},
  year      = {2022},
  pages     = {11279--11298},
  publisher = {Association for Computational Linguistics},
  url       = {https://arxiv.org/abs/2203.07540}
}

@article{oord2018cpc,
  author    = {Aaron van den Oord and Yazhe Li and Oriol Vinyals},
  title     = {Representation Learning with Contrastive Predictive Coding},
  journal   = {arXiv preprint arXiv:1807.03748},
  year      = {2018},
  url       = {https://arxiv.org/abs/1807.03748}
}

@inproceedings{yao2022react,
  author    = {Shunyu Yao and Jeffrey Zhao and Dian Yu and Nan Du and Izhak Shafran and Karthik Narasimhan and Yuan Cao},
  title     = {{ReAct}: Synergizing Reasoning and Acting in Language Models},
  booktitle = {Proceedings of the International Conference on Learning Representations (ICLR)},
  year      = {2023},
  url       = {https://arxiv.org/abs/2210.03629}
}

@inproceedings{shinn2023reflexion,
  author    = {Noah Shinn and Federico Cassano and Ashwin Gopinath and Karthik Narasimhan and Shunyu Yao},
  title     = {Reflexion: Language Agents with Verbal Reinforcement Learning},
  booktitle = {Advances in Neural Information Processing Systems (NeurIPS)},
  year      = {2023},
  volume    = {36},
  url       = {https://arxiv.org/abs/2303.11366}
}

@inproceedings{lin2023swiftsage,
  author    = {Bill Yuchen Lin and Yicheng Fu and Karina Yang and Faeze Brahman and Shiyu Huang and Chandra Bhagavatula and Prithviraj Ammanabrolu and Yejin Choi and Xiang Ren},
  title     = {{SwiftSage}: A Generative Agent with Fast and Slow Thinking for Complex Interactive Tasks},
  booktitle = {Advances in Neural Information Processing Systems (NeurIPS)},
  year      = {2023},
  volume    = {36},
  url       = {https://arxiv.org/abs/2305.17390}
}

@inproceedings{rafailov2023dpo,
  author    = {Rafael Rafailov and Archit Sharma and Eric Mitchell and Christopher D. Manning and Stefano Ermon and Chelsea Finn},
  title     = {Direct Preference Optimization: Your Language Model is Secretly a Reward Model},
  booktitle = {Advances in Neural Information Processing Systems (NeurIPS)},
  year      = {2023},
  volume    = {36},
  url       = {https://arxiv.org/abs/2305.18290}
}

@inproceedings{song2024eto,
  author    = {Chan Hee Song and Jiaman Wu and Clayton Washington and Brian M. Sadler and Weimin Shi and Yu Su},
  title     = {Trial and Error: Exploration-Based Trajectory Optimization for {LLM} Agents},
  booktitle = {Proceedings of the 62nd Annual Meeting of the Association for Computational Linguistics (ACL)},
  year      = {2024},
  url       = {https://arxiv.org/abs/2403.02502}
}

@article{jung2025co,
  title={Co-Evolving Agents: Learning from Failures as Hard Negatives},
  author={Jung, Yeonsung and Padhi, Trilok and Shaham, Sina and Khullar, Dipika and Jeong, Joonhyun and Mehrabi, Ninareh and Yang, Eunho},
  journal={arXiv preprint arXiv:2511.22254},
  year={2025}
}

@inproceedings{saycan2022,
  author    = {Michael Ahn and Anthony Brohan and Noah Brown and Yevgen Chebotar and Omar Cortes and Byron David and Chelsea Finn and Chuyuan Fu and Keerthana Gopalakrishnan and Karol Hausman and others},
  title     = {Do As I Can, Not As I Say: Grounding Language in Robotic Affordances},
  booktitle = {Conference on Robot Learning (CoRL)},
  year      = {2022},
  url       = {https://arxiv.org/abs/2204.01691}
}

@inproceedings{robinson2021hardneg,
  author    = {Joshua Robinson and Ching-Yao Chuang and Suvrit Sra and Stefanie Jegelka},
  title     = {Contrastive Learning with Hard Negative Samples},
  booktitle = {Proceedings of the International Conference on Learning Representations (ICLR)},
  year      = {2021},
  url       = {https://arxiv.org/abs/2010.04592}
}

@inproceedings{hu2022lora,
  author    = {Edward J. Hu and Yelong Shen and Phillip Wallis and Zeyuan Allen-Zhu and Yuanzhi Li and Shean Wang and Lu Wang and Weizhu Chen},
  title     = {{LoRA}: Low-Rank Adaptation of Large Language Models},
  booktitle = {Proceedings of the International Conference on Learning Representations (ICLR)},
  year      = {2022},
  url       = {https://arxiv.org/abs/2106.09685}
}

@article{qwen2025,
  author    = {{Qwen Team}},
  title     = {{Qwen2.5} Technical Report},
  journal   = {arXiv preprint arXiv:2412.15115},
  year      = {2025},
  url       = {https://arxiv.org/abs/2412.15115}
}

@article{dai2025supervised,
  title={Supervised Fine-Tuning or Contrastive Learning? Towards Better Multimodal LLM Reranking},
  author={Dai, Ziqi and Zhang, Xin and Li, Mingxin and Zhang, Yanzhao and Long, Dingkun and Xie, Pengjun and Zhang, Meishan and Li, Wenjie and Zhang, Min},
  journal={arXiv preprint arXiv:2510.14824},
  year={2025}
}

@inproceedings{he2020drrn,
  author    = {Ji He and Jianshu Chen and Xiaodong He and Jianfeng Gao and Lihong Li and Li Deng and Mari Ostendorf},
  title     = {Deep Reinforcement Learning with a Natural Language Action Space},
  booktitle = {Proceedings of the 54th Annual Meeting of the Association for Computational Linguistics (ACL)},
  year      = {2016},
  url       = {https://arxiv.org/abs/1511.04636}
}

\end{document}